\documentclass{styles/svproc}
\usepackage{url}

 \usepackage{graphicx}
 \usepackage{booktabs}%
 \usepackage[utf8]{inputenc}
\usepackage[T1]{fontenc}
\usepackage[arabic,english]{babel}
\usepackage{array}
\usepackage{booktabs}

\begin{document}
\mainmatter              
\title{Severity-Aware Weighted Loss for Arabic Medical Text Generation}
\titlerunning{Severity-Aware Weighted Loss for Arabic Medical Text Generation}  
%

\author{
Ahmed Alansary\inst{1} \and
Molham Mohamed\inst{1} \and
Ali Hamdi\inst{1}
}

\institute{
October University for Modern Sciences \& Arts, Egypt\\
\email{ahmed.mohamed406@msa.edu.eg, molham.mohamed@msa.edu.eg, ahamdi@msa.edu.eg}
}

\maketitle            

\begin{abstract}
Large language models have shown strong potential for Arabic medical text generation; however, traditional fine-tuning objectives treat all medical cases uniformly, ignoring  differences in clinical severity. This limitation is particularly critical in healthcare settings, where errors in severe cases contain higher clinical risk. In this work, we propose a severity-aware weighted loss for fine-tuning Arabic language models on medical complaint–response data. The method depends on soft severity probabilities to dynamically scale token-level loss contributions during optimization, thereby prioritizing clinically critical interactions without modifying model architectures. Experiments are conducted using the MAQA dataset, which provides Arabic medical complaints and trusted human responses. Severity labels and probabilistic scores are automatically derived using a fine-tuned AraBERT-based classifier and incorporated exclusively at the loss level. The proposed approach is evaluated across ten Arabic large language models of varying architectures and parameter scales. While standard cross-entropy fine-tuning yields only modest improvements, severity-aware optimization consistently achieves larger gains. Using a balanced weighting configuration, performance improves from 54.04\% to 66.14\% for AraGPT2-Base, from 59.16\% to 67.18\% for AraGPT2-Medium, and from 57.83\% to 66.86\% for Qwen2.5-0.5B, with peak performance reaching 67.18\%. Overall, severity-aware fine-tuning delivers improvements of up to 12.10\% over non-fine-tuned baselines, demonstrating robust and architecture-consistent gains.
\keywords{Arabic large language models, Medical text generation, Severity-aware learning, Weighted loss, Cost-sensitive optimization.}
\end{abstract}
\section{Introduction}
The rapid rise of social telehealth platforms has transformed healthcare by allowing patients to describe symptoms and seek medical guidance remotely. These platforms generate large volumes of user-generated medical text, particularly in Arabic, where patients provide detailed accounts of symptoms, disease progression, and perceived severity. While valuable for automated disease prediction and clinical decision support, this data is noisy, unstructured, and linguistically diverse, challenging traditional natural language processing (NLP) systems \cite{ICICT25,AICCSA}.

Large language models (LLMs), including BERT-based architectures, GPT variants, and LLAMA-family models, have shown strong capabilities in modeling complex medical language and enhancing tasks such as disease classification and symptom severity assessment \cite{ICICT25,ArabicLLMsMedical}. In Arabic healthcare contexts, pre-trained models like CAMeL-BERT, AraBERT, and Asafaya-BERT perform well when fine-tuned on domain-specific data \cite{ArabicSymptomClassification}. Nevertheless, domain-specific terminology, variable symptom severity, and imbalanced classes remain significant obstacles.

LLM-assisted preprocessing pipelines—using text refinement, summarization, and Named Entity Recognition (NER)—improve data quality and classification accuracy by reducing noise and emphasizing clinically relevant information \cite{ICICT25,AICCSA}. Yet, these methods rely on standard fine-tuning objectives that treat all tokens and samples equally, ignoring differences in clinical severity.

Task-specific inductive biases are particularly important in healthcare, where misclassification carries unequal risk \cite{IMSA}. Symptom severity reflects a graded spectrum, and errors involving severe cases are more critical than those involving mild symptoms. Most Arabic medical NLP frameworks, however, still optimize traditional loss functions without incorporating severity-aware learning.

To address these limitations, this work investigates severity-aware fine-tuning strategies for Arabic LLMs by integrating probabilistic severity signals into the training objective. Rather than relying solely on preprocessing or post-hoc classification, we introduce a soft severity-weighted loss that dynamically scales token-level contributions according to severity probabilities, prioritizing clinically critical information during optimization.

By combining Arabic pre-trained models with severity-informed optimization, this study advances medical text modeling beyond standard fine-tuning pipelines. The approach complements existing LLM-based preprocessing frameworks and supports more reliable, clinically aware symptom analysis in social telehealth environments.

\section{Related Work}
Recent research in Arabic natural language processing has advanced significantly with transformer-based architectures and large language models (LLMs), particularly in high-stakes domains such as healthcare. Arabic LLMs perform strongly in medical text understanding and generation when adapted to domain-specific data \cite{Ouali2025ArabicAIReview,Nazi2024LLMHealthcare,Klila2024BiomedicalLLM}.

Transformer models fine-tuned for Arabic symptom classification and diagnosis achieve notable gains through contextual modeling and data augmentation \cite{ArabicSymptomClassification,IMSA}, highlighting the importance of task-aware training for clinical decision support. Large-scale pre-trained models further enhance Arabic medical text generation under domain adaptation \cite{ArabicLLMsMedical}, and domain-specific representations trained on healthcare corpora capture clinical terminology more effectively than general-purpose embeddings \cite{Habib2021AltibbiVec}.

Reliability and safety are central concerns in medical text generation. Neural generation in regulatory and clinical contexts requires controlled mechanisms to reduce harmful or misleading outputs \cite{meyer2023neural}. Empirical studies report risks of hallucination, over-confidence, and uncontrolled generative behaviour in medical LLMs \cite{nature}, with high confidence scores not necessarily implying clinical reliability \cite{badawi2025trust}. Miscalibration therefore remains a critical challenge for safe deployment.

Parallel research has explored Arabic language understanding, generation, and alignment across multiple domains \cite{AICCSA,ICICT25,IMSA}. These studies identify challenges such as data sparsity, and dialectal variation, while demonstrating that task-specific fine-tuning consistently improves performance across classification, generation, and question-answering tasks.

Data-centric strategies have also been investigated to enhance Arabic medical NLP. Text normalization, LLM-based data and augmentation, improve robustness and generalization when annotated data are limited \cite{ArabicSymptomClassification,ICICT25}. However, these methods primarily refine input representations rather than modifying the optimization objective.

From an optimization perspective, loss-function design is critical to learning dynamics. Cross-entropy, though standard for language modeling, has limitations: vanishing gradients, overconfidence, and inadequate handling of heterogeneous error severity. Robustness can be improved via gradient approximation and confidence-sensitive adaptive formulations \cite{ApproxGradCE}, as well as linearly adaptive cross-entropy, which introduces probability-dependent adjustments to enhance efficiency while remaining computationally simple \cite{nature}.

Recent theoretical work has analyzed cross-entropy within the broader family of comp-sum losses \cite{Mao2023CE}, including generalized cross-entropy and related formulations. Non-asymptotic $H$-consistency bounds are established, upper bounding zero-one loss in terms of surrogate estimation error for a given hypothesis class. These bounds are tight and depend on quantities termed minimizability gaps.

Collectively, these findings indicate that modifying the optimization objective is a principled mechanism for improving robustness, diversity, and reliability in neural text generation. Nevertheless, limited work has explored integrating soft clinical risk signals directly into token-level LLM optimization for Arabic medical text. This gap motivates the present study, which introduces severity-aware weighted fine-tuning for Arabic medical LLMs.

\section{Dataset: Severity Augmentation}
In this study, we use the MAQA dataset, an Arabic medical question-answer corpus designed particularly for natural language processing in healthcare contexts. Each MAQA data instance represents an Arabic medical complaint expressed by a patient and a trusted answer provided by a healthcare professional. It reflects real-life conversation and covers both Modern Standard Arabic and colloquial language use.

For the experiments conducted in this study, a subset of 32,000 complaint–response pairs was selected from the full MAQA dataset for severity augmentation and model fine-tuning.

\subsection{Severity Annotation Augmentation}

The original MAQA dataset does not provide explicit severity labels or probabilistic severity information. To support risk-sensitive modeling, severity annotations are automatically derived using a fine-tuned AraBERT-based classification model applied to the complaint text. The classifier achieves an accuracy of approximately 52\%, reflecting the inherent difficulty and subjectivity of severity prediction in medical text.

For each complaint, the classifier produces a probability distribution over three predefined severity categories: \emph{non-critical}, \emph{neutral}, and \emph{critical}. The categorical severity label corresponds to the class with the highest predicted probability, while the full probability vector is retained for downstream optimization. Rather than relying solely on discrete predictions, the proposed framework leverages the full probability distribution, mitigating the impact of classification noise.

Across the augmented dataset, the severity distribution is as follows: 46\% non-critical, 30\% neutral, and 24\% critical cases. This distribution reflects a moderate class imbalance, where critical cases are underrepresented, motivating the need for severity-aware weighting during training.

\begin{table}
\centering
\caption{Example instances from the severity-aware MAQA dataset in English translations.}
\label{tab:dataset_examples}
\scriptsize
\begin{tabular}{p{0.28\textwidth} p{0.30\textwidth} p{0.12\textwidth} c c c}
\toprule
\textbf{Question} & \textbf{Answer} & \textbf{Severity} & \textbf{Non-Critical} & \textbf{Neutral} & \textbf{Critical} \\
\midrule

\foreignlanguage{arabic}{ورم في الرقبة كيف أتعامل معه هل يستدعي جراحة} &
\foreignlanguage{arabic}{على حسب مكانه ونوعه، راجع طبيب أورام} &
\foreignlanguage{arabic}{حرج} & 32\% & 32\% & 36\% \\

A neck lump, how should I deal with it? Does it require surgery? &
It depends on its location and type; consult an oncologist. &
Critical & 32\% & 32\% & 36\% \\

\midrule

\foreignlanguage{arabic}{أعاني من انتفاخ الخد نتيجة تورم اللثة بسبب تسوس الأسنان الأمامية} &
\foreignlanguage{arabic}{يحتاج الأمر إلى مضاد حيوي وعلاج عصب} &
\foreignlanguage{arabic}{غير حرج} & 37\% & 35\% & 28\% \\

I suffer from cheek swelling due to gum inflammation caused by front tooth decay. &
The condition requires antibiotic treatment and a root canal procedure. &
Non-Critical & 37\% & 35\% & 28\% \\

\midrule

\foreignlanguage{arabic}{هل الكركم يساهم في ارتفاع ضغط الدم وما هي فوائده وكيفية استعماله} &
\foreignlanguage{arabic}{على العكس، كما هو معروف، يساعد على خفض الضغط} &
\foreignlanguage{arabic}{متوسط} & 35\% & 36\% & 29\% \\

Does turmeric contribute to high blood pressure? What are its benefits and how is it used? &
On the contrary, it helps lower blood pressure. &
Neutral & 35\% & 36\% & 29\% \\

\bottomrule
\end{tabular}
\end{table}

\subsection{Final Dataset Structure}

After augmentation, each instance contains six fields:

\begin{itemize}
    \item \textbf{Question}: Arabic medical complaint in free-form text.
    \item \textbf{Answer}: Trusted human-generated medical response.
    \item \textbf{Severity}: Predicted categorical severity label.
    \item \textbf{Non-Critical}: Probability of the complaint being non-critical.
    \item \textbf{Neutral}: Probability of the complaint being neutral.
    \item \textbf{Critical}: Probability of the complaint being critical.
\end{itemize}

Table~\ref{tab:dataset_examples} presents representative examples illustrating these fields and their associated severity distributions. The probabilistic severity scores provide a continuous representation of clinical risk, enabling more flexible modeling compared to hard labels alone.

\section{Methodology}
A severity-aware fine-tuning framework is proposed for Arabic medical text generation, incorporating clinical severity into the optimization objective. It modulates each training instance at the token level using soft severity probabilities, aligning gradient updates with medical risk sensitivity.

\begin{figure}[h]
\centering
\includegraphics[width=0.3\columnwidth]{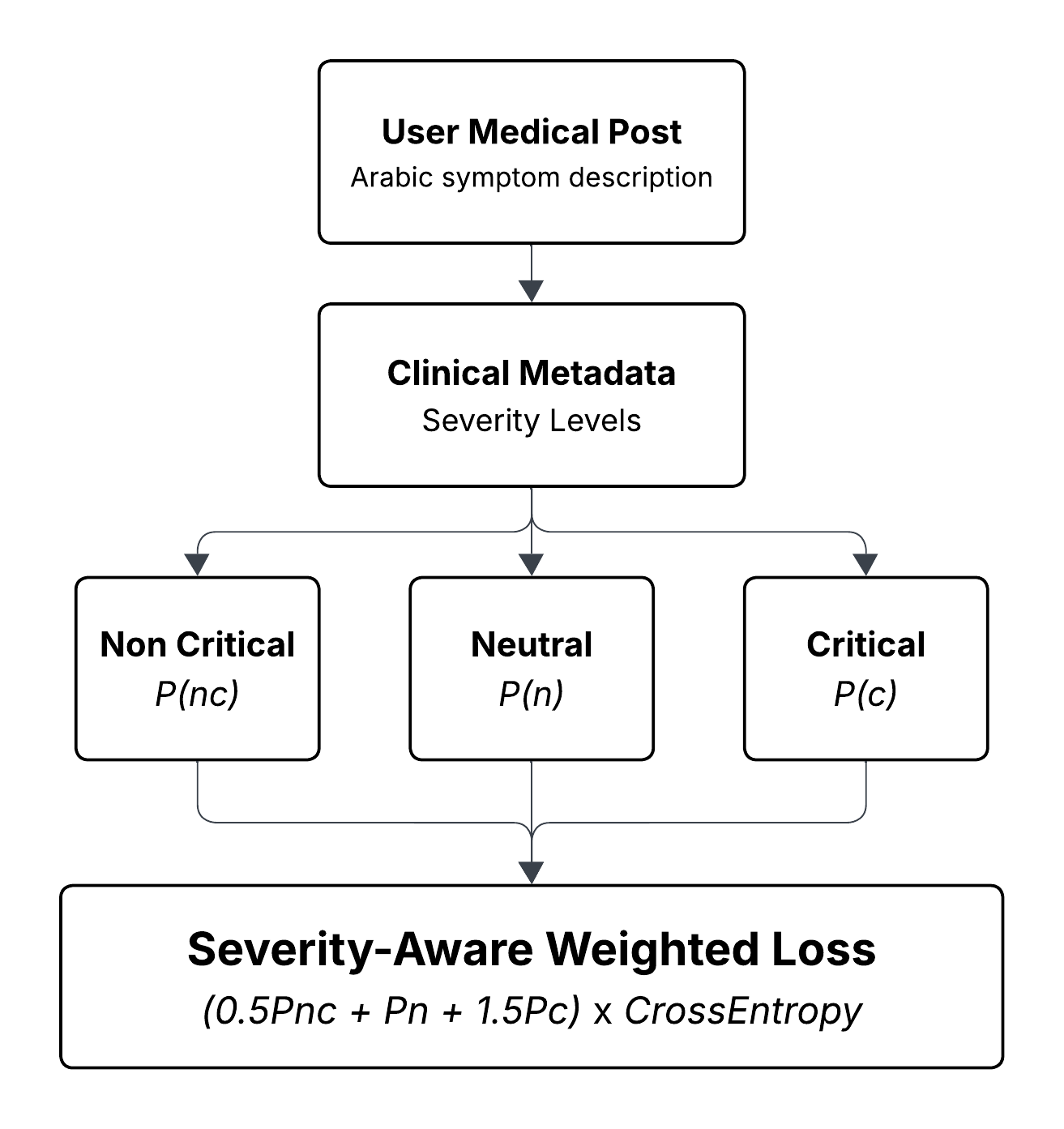}
\caption{Overview of the proposed severity-aware training process.}
\label{fig:loss_process}
\end{figure}

Figure~\ref{fig:loss_process} illustrates the overall training pipeline, highlighting how severity probabilities are integrated into the loss computation during model fine-tuning.

\subsection{Problem Formulation}
Each dataset instance consists of an input sequence $X = (x_1, \dots, x_T)$ representing a medical complaint, a target sequence $Y = (y_1, \dots, y_L)$ with a trusted human-generated response, and a soft severity distribution $\mathbf{s} = [p_{nc}, p_{n}, p_{c}]$ for non-critical, neutral, and critical probabilities.

Severity probabilities, derived from a fine-tuned AraBERT classifier, serve as auxiliary supervision during fine-tuning. The goal is to adapt a pre-trained Arabic LLM to generate medically appropriate responses while prioritizing clinically critical cases.

\subsection{Baseline Language Modeling Objective}

Standard auto-regressive fine-tuning of LLMs relies on the cross-entropy loss:

\[
\mathcal{L}_{CE} = - \sum_{t=1}^{L} \log P(y_t \mid y_{<t}, X),
\]

where $P(y_t \mid y_{<t}, X)$ denotes the predicted probability of token $y_t$ conditioned on the input sequence and previously generated tokens. This objective treats all instances uniformly, regardless of their associated clinical severity.

\subsection{Severity-Aware Weighted Loss}

To incorporate risk sensitivity, a severity-aware weighted loss is introduced. For each instance, a continuous weight $w$ is computed as a linear combination of the severity probabilities:

\[
w = \alpha \, p_{nc} + \beta \, p_{n} + \gamma \, p_{c},
\]

where $\alpha$, $\beta$, and $\gamma$ are scalar coefficients controlling the relative emphasis on non-critical, neutral, and critical cases, respectively.

The severity-aware objective becomes:

\[
\mathcal{L}_{SA} = \frac{1}{L} \sum_{t=1}^{L} w \cdot \left( - \log P(y_t \mid y_{<t}, X) \right).
\]

This formulation amplifies gradient contributions from clinically critical samples while preserving learning from lower-risk cases.

\subsection{Gradient and optimization Dynamics}

The weighting coefficient $w$ scales token-level gradients during backpropagation. Each token's gradient is multiplied by $w$, producing larger updates for instances with higher severity. Larger $\gamma$ values amplify the learning signal for critical samples, accelerating their effect on parameter updates.

Smaller $\alpha$ values down-weight non-critical instances, prioritizing high-risk cases while retaining exposure to lower-risk data. Thus, $(\alpha, \beta, \gamma)$ controls the trade-off between stability and risk sensitivity during training.

\subsection{Weighting Configurations}

To analyze the impact of risk emphasis, three weighting configurations are investigated:

\begin{itemize}
    \item \textbf{Mild setting:} $\alpha=0.75, \beta=1, \gamma=1.25$
    \item \textbf{Strong setting:} $\alpha=0.25, \beta=1, \gamma=1.75$
    \item \textbf{Balanced setting:} $\alpha=0.5, \beta=1, \gamma=1.5$
\end{itemize}

These configurations progressively increase the influence of critical complaints during optimization. The comparative evaluation of these settings enables systematic analysis of how varying degrees of risk emphasis affect model performance.

\subsection{Training Procedure}

During fine-tuning, each complaint--response pair is tokenized and processed using a pre-trained Arabic LLM. The model produces token-level logits, which are transformed into log-probabilities via a softmax operation. The negative log-likelihood is computed per token and subsequently scaled by the instance-specific severity weight before aggregation.

Severity information is incorporated exclusively at the loss level and is not appended to the model input. This design avoids altering language representations while enabling risk-aware gradient modulation.

As illustrated in Figure~\ref{fig:loss_process}, severity weighting is applied after token-level loss computation and prior to final loss aggregation.

\subsection{Implementation Details}

The severity-aware loss is implemented by extending the standard training loop to accept automatically inferred severity probabilities alongside textual inputs. Token-level losses are computed without reduction, reweighted using the corresponding severity coefficient, and averaged to obtain the final training loss. The method is fully compatible with existing transformer-based fine-tuning pipelines and introduces minimal computational overhead.

\section{Results}
The proposed severity-aware weighted loss is evaluated on ten Arabic LLMs with diverse parameter scales, pre-training corpora, and linguistic coverage. Table~\ref{tab:extended_results} compares three configurations: (i) the original pre-trained models (Base), (ii) standard cross-entropy fine-tuning (CE), and (iii) three severity-aware weighting strategies varying in emphasis on clinically critical cases.

\subsection{Baseline vs. Standard Fine-Tuning}

Standard cross-entropy fine-tuning produces modest yet consistent improvements over the Base models. Although task adaptation enhances performance for most LLMs, the gains remain limited in several cases, indicating that uniform token-level optimization does not sufficiently account for heterogeneous clinical risk within the dataset.

\begin{table*}[t]
\centering
\footnotesize
\renewcommand{\arraystretch}{0.5}
\caption{Performance comparison across baseline, standard cross-entropy fine-tuning, and severity-aware weighting configurations.}
\label{tab:extended_results}
\begin{tabular}{l c c c c c}
\toprule
\textbf{Model} 
& {\textbf{Base}} 
& {\textbf{Cross-Entropy}} 
& {\textbf{SA (Mild)}} 
& {\textbf{SA (Strong)}} 
& {\textbf{SA (Balanced)}} \\
\midrule
AraGPT2-Base        & 54.04\% & 59.71\% & 59.75\% & 65.91\% & \textbf{66.14\%} \\
AraGPT2-Medium      & 59.16\% & 60.14\% & 60.48\% & 60.55\% & \textbf{67.18\%} \\
Bloomz-560M         & 63.59\% & 63.65\% & 63.91\% & 63.79\% & \textbf{65.52\%} \\
Al-Atlas-0.5B       & 61.03\% & 61.52\% & 62.03\% & 65.02\% & \textbf{66.83\%} \\
GPT2-Small-Arabic   & 56.40\% & 58.88\% & 62.01\% & 60.92\% & \textbf{63.06\%} \\
LFM2-350M           & 63.34\% & 63.72\% & 65.24\% & 64.08\% & \textbf{66.60\%} \\
LFM2-700M           & 63.46\% & 63.88\% & 64.63\% & 64.25\% & \textbf{67.18\%} \\
Dialect-ar-gpt-2021 & 60.41\% & 60.73\% & 62.12\% & 60.52\% & \textbf{67.13\%} \\
Qwen2.5-0.5B        & 57.83\% & 57.96\% & 58.68\% & 65.15\% & \textbf{66.86\%} \\
Qwen3-0.6B          & 60.43\% & 61.40\% & 62.87\% & 62.53\% & \textbf{65.06\%} \\
\bottomrule
\end{tabular}
\end{table*}

\subsection{Impact of Severity-Aware Weighting}

Severity-aware loss modulation consistently outperforms standard cross-entropy fine-tuning. Across all evaluated LLMs, at least one severity-aware configuration exceeds the CE baseline, demonstrating the benefit of incorporating risk-sensitive signals into the optimization.

The three weighting strategies show distinct behaviors:

\textbf{Mild Setting:}
\{$0.75P(NC) + 1P(N) + 1.25P(C)$\}\\
This configuration lightly prioritizes critical cases, yielding steady but moderate gains over CE for several LLMs (e.g., LFM2-350M, Qwen3-0.6B). Overall improvements remain limited, suggesting mild weighting does not sufficiently amplify clinically critical samples.\\

\textbf{Strong Setting:}
\{$0.25P(NC) + 1P(N) + 1.75P(C)$\}\\
Heavily prioritizing critical complaints, this setting produces substantial gains for some LLMs (e.g., AraGPT2-Base, Qwen2.5-0.5B) but can reduce stability across architectures. Aggressive down-weighting of non-critical samples may slightly lower overall performance, indicating potential optimization imbalance.\\

\textbf{Balanced Setting:}
\{$0.5P(NC) + 1P(N) + 1.5P(C)$\}\\
The balanced configuration achieves the most consistent and highest performance across most models, including AraGPT2 variants, Al-Atlas-0.5B, LFM2-350M, LFM2-700M, Dialect-ar-gpt-2021, Qwen2.5-0.5B, and Qwen3-0.6B. Moderate emphasis on critical cases provides an effective trade-off between clinical risk sensitivity and generalization.

\subsection{Model-Level Observations}

Severity-aware optimization benefits LLMs of all sizes. Smaller and mid-sized models, such as AraGPT2 variants and Qwen2.5-0.5B, gain most, as risk-sensitive weighting directs gradients toward clinically important samples when capacity is limited.

Larger models, including LFM2-350M, LFM2-700M, and Bloomz-560M, also improve, though with smaller margins, showing the method enhances performance even with strong representations.

Gains hold across Modern Standard Arabic and dialectal data, with Dialect-ar-gpt-2021 demonstrating generalization across linguistic varieties.

\subsection{Overall Analysis}

The empirical findings confirm that integrating soft severity distributions directly into the loss function provides a more effective optimization strategy than uniform cross-entropy fine-tuning for Arabic medical LLMs. While different weighting configurations influence optimization dynamics differently, the balanced setting achieves the most reliable and architecture-consistent improvements.

Risk-sensitive loss modulation therefore emerges as a practical and computationally lightweight approach for enhancing Arabic medical text generation without modifying model architectures or increasing training complexity.

\section{Conclusion}
This paper introduced a severity-aware weighted loss for fine-tuning Arabic large language models (LLMs) for medical text generation. Unlike traditional objectives that treat all instances uniformly, the proposed approach incorporates soft severity probabilities into the optimization, assigning greater weight to clinically critical interactions.

Empirical evaluation across ten Arabic LLMs shows that standard cross-entropy fine-tuning yields modest gains, whereas severity-aware optimization consistently delivers larger improvements across architectures, parameter scales, and linguistic varieties. These results indicate that uniform token-level optimization fails to capture heterogeneous clinical risk in medical dialogue data.

An ablation of three weighting strategies reveals distinct behaviors. While mild and strong settings improve performance in specific cases, the balanced configuration \{$0.5P(NC) + 1P(N) + 1.5P(C)$\} achieves the most stable and consistently superior results, providing an effective trade-off between risk sensitivity and generalization.

Severity labels and probabilities, derived from a fine-tuned AraBERT classifier, enable severity-aware learning without additional manual annotation. Integrating severity solely at the loss level keeps the framework computationally lightweight while aligning gradient updates with clinical importance.

Despite these strengths, the approach relies on automatically inferred severity signals and fixed weighting coefficients, which may introduce noise and calibration limitations. Future work could explore adaptive weighting, joint modeling of severity and generation, and more comprehensive evaluation protocols.

Overall, severity-aware loss modulation offers a practical and scalable strategy for enhancing Arabic medical LLMs toward safer, clinically aligned text generation.

\bibliographystyle{splncs04}
\bibliography{templates/ref}

\end{document}